\documentclass[preprint,12pt,authoryear]{elsarticle}

\usepackage{amsmath,amsfonts}
\usepackage{algorithmic}
\usepackage{algorithm}
\usepackage{array}
 \usepackage{subcaption}
\usepackage{textcomp}
\usepackage{stfloats}
\usepackage{subcaption}
\usepackage{threeparttable} 
\usepackage{url}
\usepackage{verbatim}
\usepackage{graphicx}
\usepackage{booktabs}
\usepackage{multirow}
\usepackage{diagbox}
\usepackage{amssymb}
\usepackage{bm}
\usepackage{pifont}
\usepackage[table]{xcolor}  
\usepackage[colorlinks,
linkcolor=blue,
anchorcolor=blue,
urlcolor = blue,
citecolor=blue]{hyperref}


\journal{Expert Systems With Applications}

\begin{document}

\begin{frontmatter}



\title{Efficient Neural Combinatorial Optimization Solver for the Min-max Heterogeneous Capacitated Vehicle Routing Problem}


\author[address_a]{Xuan Wu} 
\author[address_b]{Di Wang}
\author[address_a]{Chunguo Wu}
\author[address_c]{Kaifang Qi} 
\author[address_b]{Chunyan Miao}   
\author[address_a]{Yubin Xiao\corref{mycorrespondingauthor}}
\author[address_a]{Jian Zhang\corref{mycorrespondingauthor}}
\author[address_a]{You Zhou\corref{mycorrespondingauthor}}
\cortext[mycorrespondingauthor]{Corresponding authors}

\address[address_a]{Key Laboratory of Symbolic Computation and Knowledge Engineering of Ministry of Education, College of Computer Science and Technology, Jilin University, Changchun, 130012, China}
\address[address_b]{Joint NTU-UBC Research Centre of Excellence in Active Living for the Elderly, Nanyang Technological University, 639798, Singapore}
\address[address_c]{College of Software, Jilin University, Changchun, 130012, China}
\begin{abstract}
Numerous Neural Combinatorial Optimization (NCO) solvers have been proposed to address Vehicle Routing Problems (VRPs). However, most of these solvers focus exclusively on single-vehicle VRP variants, overlooking the more realistic Min-Max Heterogeneous Capacitated Vehicle Routing Problem (MMHCVRP), which involves multiple vehicles. Existing MMHCVRP solvers typically select a vehicle and its next node to visit at each decoding step, but often make myopic  decisions during decoding and overlook key properties of MMHCVRP, including local topological relationships, vehicle permutation invariance, and node symmetry, resulting in suboptimal performance. To better address these limitations, we propose ECHO, an efficient NCO solver. First, ECHO exploits the proposed dual-modality node encoder to capture local topological relationships among nodes. Subsequently, to mitigate myopic decisions, ECHO employs the proposed Parameter-Free Cross-Attention mechanism to prioritize the vehicle selected in the preceding decoding step. Finally, leveraging vehicle permutation invariance and node symmetry, we introduce a tailored data augment strategy for MMHCVRP to stabilize the Reinforcement Learning training process. To assess the performance of ECHO, we conduct extensive experiments. The experimental results demonstrate that ECHO outperforms state-of-the-art (SOTA) NCO solvers across varying numbers of vehicles and nodes, and exhibits well-performing generalization across both scales and distribution patterns. Finally, ablation studies validate the effectiveness of all proposed methods. 
\end{abstract}


\end{frontmatter}



\section{Introduction}
\label{sec1}
 
Vehicle Routing Problem (VRP) is a fundamental class of Combinatorial Optimization Problem (COP) with widespread applications in diverse domains, e.g., communication and transportation \citep{airport_tits, bengio_machine_2021, xiao2023reinforcement,wu2025,9115005}. Conventional algorithms for solving VRPs can be broadly divided into three categories, namely exact, approximation, and heuristic algorithms \citep{li_research_2021,wu_incorporating_2023,ugas,wu_neural_2023}. However, these conventional algorithms cannot derive insights from historical VRP instances and lack the capability to solve multiple VRP instances in parallel, thus leading to significant computing overhead \citep{wu_2024_survey}.

Recently, numerous Neural Combinatorial Optimization (NCO) solvers have been proposed to address the VRP \citep{xiao2024improving, 9901466, LI2024124514,xiao_distilling_2024,LUO2025127311,liyuanshu}. These solvers can derive insights from historical instances and exploit GPU parallelism to process multiple VRP instances concurrently, achieving high-level performance with substantially reduced computational time \citep{xin_multi-decoder_2021, glop}. However, most existing NCO solvers focus on either the unconstrained Traveling Salesman Problem (TSP) or the Capacitated Vehicle Routing Problem (CVRP) with only basic capacity constraints \citep{mingzhao, mvmoe}. In practical applications, VRP variants often involve more complex attributes or constraints, thereby hindering the deployment of existing NCO research \citep{li2024cada,drl}. For example,  real-world scenarios frequently feature vehicles with heterogeneous attributes (e.g., capacity and speed), and fleet operators commonly aim to minimize the longest travel time among all vehicles  (min-max) rather than the total time traveled by the entire fleet (min-sum). This problem is known as the Min-Max Heterogeneous Capacitated Vehicle Routing Problem (MMHCVRP).

Existing NCO methods for the MMHCVRP are typically trained via Reinforcement Learning (RL) and can be broadly categorized into two categories, namely AutoRegressive (AR) and Parallel-Autoregressive (PAR) solvers. AR-based solvers adopt an encoder-decoder architecture, where vehicle and node encoders embed vehicle and node features, respectively, and the decoder sequentially selects a vehicle and its next node to visit at each time step \citep{drl,2dptr}. PAR-based solvers similarly employ the node and vehicle encoders to embed their respective features. However, during decoding, each vehicle simultaneously selects a node to visit at each time step. This parallel selection enables PAR-based solvers to achieve faster inference than their AR-based counterparts. However, the resulting high‑dimensional joint action space complicates training and may produce suboptimal solutions \citep{dpn}. Moreover, conflicts arise when multiple vehicles simultaneously select the same node within a single time step. AR-based solvers also exhibit certain limitations. As shown in Figure~\ref{fig_comparison}(a), the SOTA AR-based solver 2D-Ptr  disregards the priority of the vehicle selected in the preceding time step and rely solely on the accumulated travel time of each vehicle for decision-making. Consequently, at the third time step, vehicle $m_3$ is chosen to visit the node $n_2$. While this short‑sighted strategy may produce plausible solutions initially, it ultimately leads to suboptimal performance. Moreover, both these AR-based and NAR-based solvers neglect key properties of the MMHCVRP, namely local topological relationships, vehicle permutation invariance, and node symmetry, which ultimately degrades their performance.

To better address these limitations, we propose an Efficient Neural Combinatorial Optimization Solver for the Min-max Heterogeneous Capacitated Vehicle Routing Problem (\textbf{ECHO}). Specifically, to better capture local topological features, we replace the vanilla node encoder with our dual-modality node encoder, which employs a cross‑attention mechanism to integrate node and edge features (see Section~\ref{sec4.1} for more details). This design endows ECHO with robust generalization capabilities across both scales and distribution patterns. In addition, to mitigate myopic decisions, we propose a Parameter-Free Cross-Attention (PFCA) that prioritizes the node selected at the preceding time step during decoding. Specifically, PFCA efficiently integrates information about the vehicle selected in the $t-1$th time step into the node embeddings. Subsequently, ECHO selects a feasible vehicle-node pair at the $t$th time step based on the updated node embeddings and current vehicle embeddings (see Section~\ref{sec4.2} for more details). As shown in Figure~\ref{fig_comparison}(b),
in the third decoding step, vehicle $m_2$ is selected to visit the node $n_2$ instead of vehicle $m_3$, ultimately yielding a superior solution. To the best of our knowledge, ECHO is the first work to explicitly model historically selected vehicle information and design a tailored mechanism to leverage it for guiding sequential decision-making, thereby effectively alleviating myopic decisions.  Finally, we propose a tailored data augment method for MMHCVRP, which effectively generates diverse instances by leveraging vehicle permutation invariance and node symmetry to stabilize the RL training process. 

To assess the effectiveness of the proposed ECHO, we conduct extensive experiments on MMHCVRP instances with different numbers of vehicles and nodes. The experimental results demonstrate that ECHO outperforms the other NCO solvers designed for MMHCVRP, including two AR-based NCO solvers, namely DRL \citep{drl} and 2D-Ptr \citep{2dptr}, as well as one PAR-based solver, namely PARCO \citep{parco}. Notably, ECHO reduces the average gap by approximately 3\% compared to the SOTA solver PARCO across all vehicle and node scales (see Section~\ref{sec5.2}). In addition, ECHO exhibits robust cross-scale and cross-distribution generalization ability. Finally, ablation studies demonstrate the effectiveness of the proposed dual-modality node encoder, PFCA mechanism, and data augment method, while the case study further illustrates the effectiveness of PFCA in avoiding the neglect of the priority of previously selected vehicles.

The key contributions of this work are as follows:
\begin{itemize} 
\item We propose a dual-modality node encoder, which effectively exploits the cross attention to fuse node and edge features, thereby capturing local topological relationships.

\item We propose a novel decoder that integrates the tailored PFCA mechanism, enabling the solver to focus more on the vehicle selected in the preceding decoding step.

\item We propose an efficient data augment method for the MMHCVRP, which exploits problem-specific symmetries to avoid local minima and stabilize the training process.

\item The experimental results demonstrate that the proposed ECHO achieves SOTA performance compared to both AR- and PAR-based NCO solvers, while also demonstrating high-level generalization ability. Moreover, ablation studies validate the effectiveness of all proposed methods.
\end{itemize} 

The remainder of this paper is organized as follows: Section~\ref{sec2}
reviews the related work in the field of NCO. Section~\ref{sec3} presents the definition of MMHCVRP and the MDP formula used in our method. Section~\ref{sec4} describes the architecture of ECHO and the proposed data augment strategy. Section~\ref{sec5} discusses the experimental results. Section~\ref{sec6} discusses the limitations of the proposed ECHO. Finally,
Section~\ref{sec7} draws the conclusion.

\section{Related Work}
\label{sec2}
In this section, we first introduce general NCO solvers for VRPs, and then discuss the existing efforts for solving the more complex MMHCVRP.
\subsection{NCO solvers for VRP}
\label{sec2.1}
Existing NCO solvers for VRP can be broadly categorized into two types, namely Learning to Construct and Learning to Improve solvers \citep{li2025destroy,wu_learning_2022,GUAN2025126961}. Inspired by sequence‑to‑sequence models in the machine translation task, Learning to Construct solvers sequentially select unvisited nodes and append them to partial solutions to construct the complete solutions \citep{vinyals_pointer_2015, kool_attention_2019, luo_neural_2023}. In contrast, Learning to Improve solvers, motivated by ruin‑and‑repair heuristics, iteratively perturb and repair current complete solutions to explore (sub-)optimal solutions within a given time frame  \citep{ma_learning_2021,ma_learning_2023,chen_learning_2019}. Among these two paradigms, the Learning‑to‑Construct solver is more prevalent.

As highlighted in the recent survey by Wu et al. \citep{wu_2024_survey}, although these solvers have achieved significant progress, they still encounter certain limitations. Specifically, existing NCO solvers exhibit poor cross‑scale and cross-distribution generalization abilities and cannot solve large-scale VRPs in real-time \citep{xiao2025, glop, luo2025boosting,dgl}. In addition, their applicability remains largely confined to TSP and CVRP, with limited success on more complex VRP variants \citep{mvmoe,li2024cada}. An important variant is the min-max VRP, where the objective is to minimize the maximum route duration across the entire vehicle fleet. Although several studies have examined the min-max TSP \citep{Liang_Splitnet_2023, 10040987, kim2023learning, dpn, et}, they overlook vehicle heterogeneity, and the simplified constraints of min-max TSP hinder applicability to real-world routing scenarios. Consequently, it is essential to investigate the MMHCVRP, which incorporates these additional practical constraints.


 \subsection{NCO solvers Specifically Designed for MMHCVRP}
\label{sec2.2}
To better solve the more challenging MMHCVRP, \cite{drl} introduced DRL, the first AR-based solver tailored for this problem. Specifically, DRL employs an attention-based encoder and a Fully Connected Network (FCN) to embed vehicle and node features, respectively. During decoding, at each time step, it employs a policy network that first selects a vehicle and then chooses that vehicle's next node to visit, iteratively constructing complete routes until all customer nodes have been visited. Building on the prior study \citep{drl}, \cite{2dptr} proposed 2D-Ptr, which incorporates a more efficient decoder. Specifically, at each time step, 2D-Ptr directly computes the dot product between vehicle and node embeddings, and selects the vehicle-node pair with the highest scores as the current vehicle and its next node to visit. Compared to DRL’s sequential first-vehicle‑then‑node selection, 2D‑Ptr's joint selection of vehicle-node pair significantly enhances the computational efficiency. However, both AR-based NCO solvers overlook critical MMHCVRP properties, namely local topological relationships, vehicle permutation invariance, and node symmetry. In addition, they disregard the priority of the previously selected vehicle during the decoding process, resulting in suboptimal solutions. To better solve these limitations, we propose ECHO (see Section~\ref{sec4} for more details).

Unlike AR methods that sequentially select one vehicle and its next node to visit at each time step, \cite{parco} proposed PARCO, a PAR-based solver for MMHCVRP that simultaneously selects each vehicle’s next node at every decoding step. Theoretically, PAR‑based solvers require only $\frac{1}{M}$ of the time steps of AR‑based solvers, where $M$ denotes the number of vehicles, thereby achieving superior efficiency. However, PARCO suffers from a high‑dimensional joint action space that complicates training and necessitates conflict handlers among vehicles. Consequently, although PARCO adopts a strong training paradigm, its performance remains constrained \citep{dpn}. In contrast, we extend the augmentation scheme used in PARCO and introduce a more specifically tailored architecture, which jointly contributes to the superior performance of ECHO.

\begin{table}[!t]
    \centering
    \small
    \caption{Notations Used in MDP}
    \label{tabparams}
    \begin{tabular}{c| c|c}
    \toprule
    Notation & Definition & Category\\ \midrule
     $\chi_i$  & Speed of $i$th vehicle & Static \\
    $\rho_i$  & Capacity of $i$th vehicle  & Static \\
     $\hat{\rho}_i^t$ & Accumulated usage capacity of $i$th vehicle at $t$th step & Dynamic \\
    $\delta^t_i$ & Accumulated travel time of $i$th vehicle at $t$th step & Dynamic \\
     $\ell_{i}^{t}$  & The last visited node of $i$th vehicle  & Dynamic \\ \midrule
    $x_j$ & X-coordinate of the $j$th node & Static \\
     $y_j$ & Y-coordinate of the $j$th node &  Static\\
    $d_j^t$ & Demand of $j$th node at $t$th step &  Dynamic\\
    \bottomrule
    \end{tabular}
\end{table}
\section{Preliminary}
\label{sec3}
In this section, we first introduce the definition of MMHCVRP, then present the Markov Decision Process (MDP) formula used in our method.
\subsection{Definition of MMHCVRP}
\label{sec3.1}
Consider an MMHCVRP instance $G(\mathbb{M},\mathbb{N})$, where $\mathbb{M} = \{m_i\}_{i=1}^{M}$ denotes a fleet of $M$ heterogeneous vehicles, and $\mathbb{N}= \{n_j\}_{j=0}^{N}$ denotes a set of nodes comprising a depot $n_0$ and the number of $N$ customers. In this problem, each fully loaded vehicle departs from the depot and visits a sequence of customers to satisfy their demands, subject to the constraints that each customer is visited exactly once and the load carried by any vehicle along its route does not exceed its capacity. Each vehicle $m_i$ is characterized by two attributes, namely, capacity $\rho_i$ and speed $\chi_i$. Each node $n_i$ is described by three attributes, namely, coordinates $x_j$ and $y_j$, as well as demand $q_j$. In addition, the objective is to minimize the maximum route duration among all vehicles, which are defined in Eq.~(\ref{eq1}).
\begin{equation}
\label{eq1}
    \min \max_{i\in \mathbb{M} } (\sum_{j=1}^{|\gamma_i|} \frac{dist(\gamma_i^{j-1}, \gamma_i^j)}{\chi_i}),
\end{equation}
where $\gamma_i$ denotes the route of the $i$th vehicle, and $\gamma_i^j$ denotes the $j$th visiting node of the $i$th vehicle. Functions $|\cdot|$ and $dist(\cdot,\cdot)$ denote the cardinality operator and distance between two sequential nodes, respectively. 

\subsection{Markov Decision Process}
\label{sec3.2}

Following the prior studies \citep{drl, 2dptr}, we adopt the AR paradigm and define an identical MDP to derive the visitation sequence $\tau = \{(m^t, n^t), t=\{0,1, ..., T-1 \} \}$ for solving the MMHCVRP. Specifically, the MDP comprises the following five components:

    
\subsubsection{\textbf{State}} The current state is defined as $s^t = (m_i^t, n_i^t), i\in \{1, \cdots, M\}, j\in\{0, \cdots, N\}$, where $m_i^t$ denotes the $i$th vehicle's state comprising of $ (\chi_i, \rho_i, \hat{\rho}_i^t, \delta^t_i, \ell_{i}^{t})$ and $n_i^t$ denotes the $j$th node's state comprising of $ (x_j, y_j, q_j^t)$. The definitions of these notations are presented in Table~\ref{tabparams}.

\subsubsection{\textbf{Action}} The action $a^t$ is defined as selecting a vehicle-node pair $(m_i^t, n_j^t)$, indicating that the $i$th vehicle is selected to visit the $j$th node at the $t$th time step. Note that, the action space comprises only valid vehicle-node combinations, while invalid pairs are masked, e.g., those involving customer nodes that have already been visited.

\begin{figure}[!t]
	\centering
\includegraphics[scale=0.75]{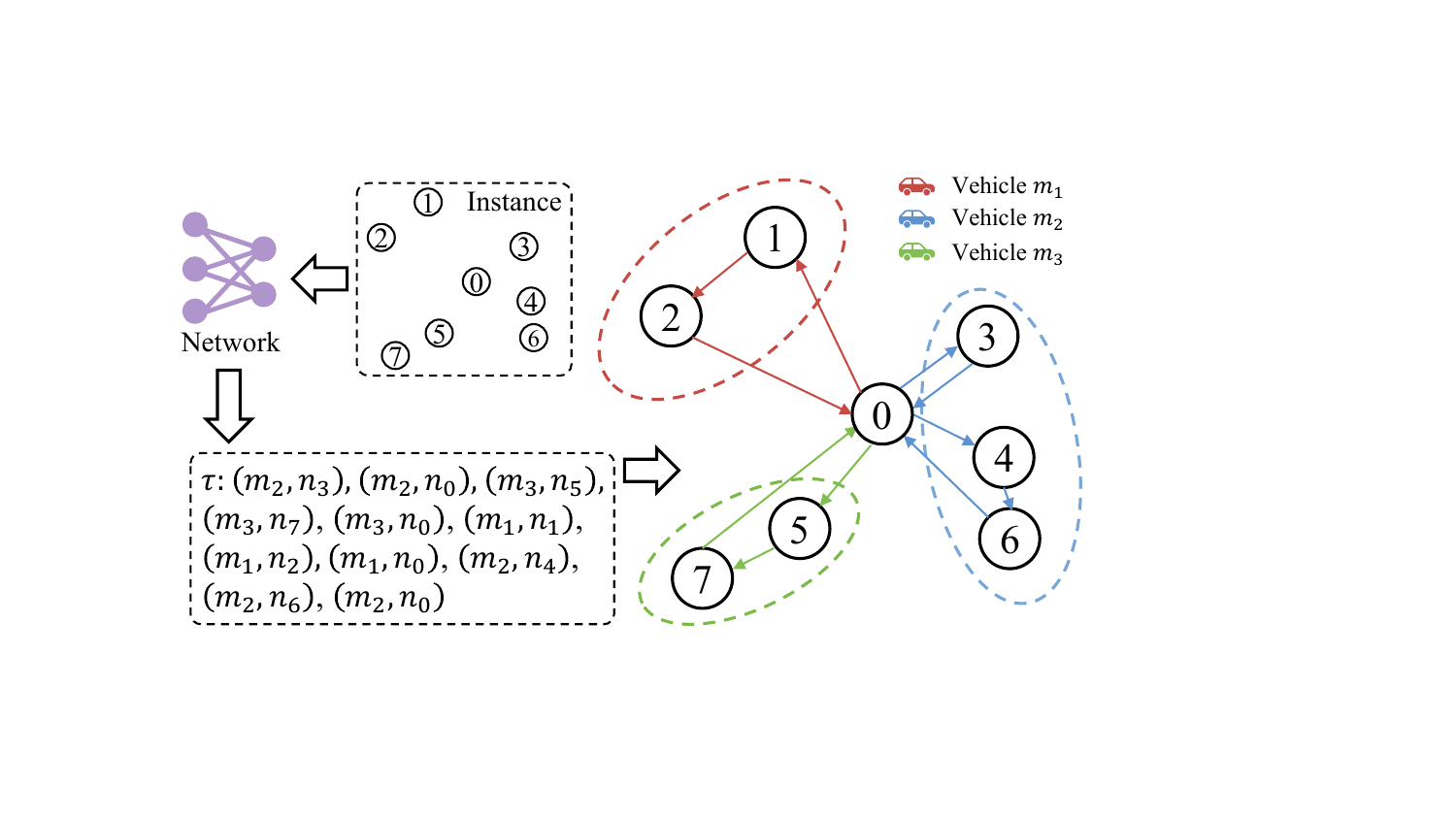}
	\caption{Illustration of solution process for an MMHCVRP instance with three vehicles and eight nodes (seven customer nodes and one depot node). At each time step, the policy network selects a vehicle and a node as the action, repeating until all customer nodes have been served and all vehicles have returned to the depot.} 
	\label{figexample}
\end{figure}
\subsubsection{\textbf{Transaction}} Given the current state $s^t$ and action $a^t = (m_i^t, n_j^t)$, the next state $s^{t+1} = f(s^t, a^t)$ is obtained by updating each vehicle's accumulated usage capacity, accumulated travel time, and current location, as well as each node's demand. For every vehicle $k \in \{1, \cdots, M\}$ and node $l \in \{0, \cdots, N\}$, these updates are defined in Eqs.~(\ref{eq2})$\sim$(\ref{eq5}).
\begin{equation}
\label{eq2}
\hat{\rho}_k^{t + 1} = 
\begin{cases} 
0, & k  = i, j=0, \\
\hat{\rho}_k^{t} + d_j, & k  = i, j\neq 0, \\
\hat{\rho}_k^{t}, & \text{otherwise.}
\end{cases}
\end{equation}
\begin{equation}
\delta_k^{t + 1} = 
\begin{cases} 
\delta_k^t + \dfrac{dist(n_j, \ell_{i}^{t})}{\chi_m}, & k = i, \\
\delta_k^t, & \text{otherwise.}
\end{cases}
\end{equation}
\begin{equation}
\ell_k^{t + 1} = 
\begin{cases} 
n_j, & k = i, \\
\ell_k^t, & \text{otherwise.}
\end{cases} 
\end{equation}
\begin{equation}
\label{eq5}
d_l^{t + 1} = 
\begin{cases} 
0, & l = j, \\
d_l^t, & \text{otherwise.}
\end{cases}
\end{equation}
\subsubsection{\textbf{Reward}} The total reward $R$ is defined as the negative of the longest travel time across all vehicles, as given in Eq.~(\ref{eq6}). 
\begin{equation}
\label{eq6}
R = -\max_{i\in \{1, \cdots, M\}} \delta^T_i,
\end{equation} 
where $\delta^T_i$ denotes the accumulated travel time of the $i$th vehicle at final step, and $M$ denotes the number of vehicles.

\subsubsection{\textbf{Policy}}
Given the current state $s_t$, the policy selects an action according to the probability produced by the neural network. This process continues until all customer nodes are visited and all vehicles return to the depot, concluding the episode and resulting in the construction of a valid tour. In Figure~\ref {figexample}, we illustrate the visiting sequence generated by the policy network for an MMHCVRP instance.

\section{ECHO}
\label{sec4}
\begin{figure*}[!t]
	\centering
	\includegraphics[scale=0.85]{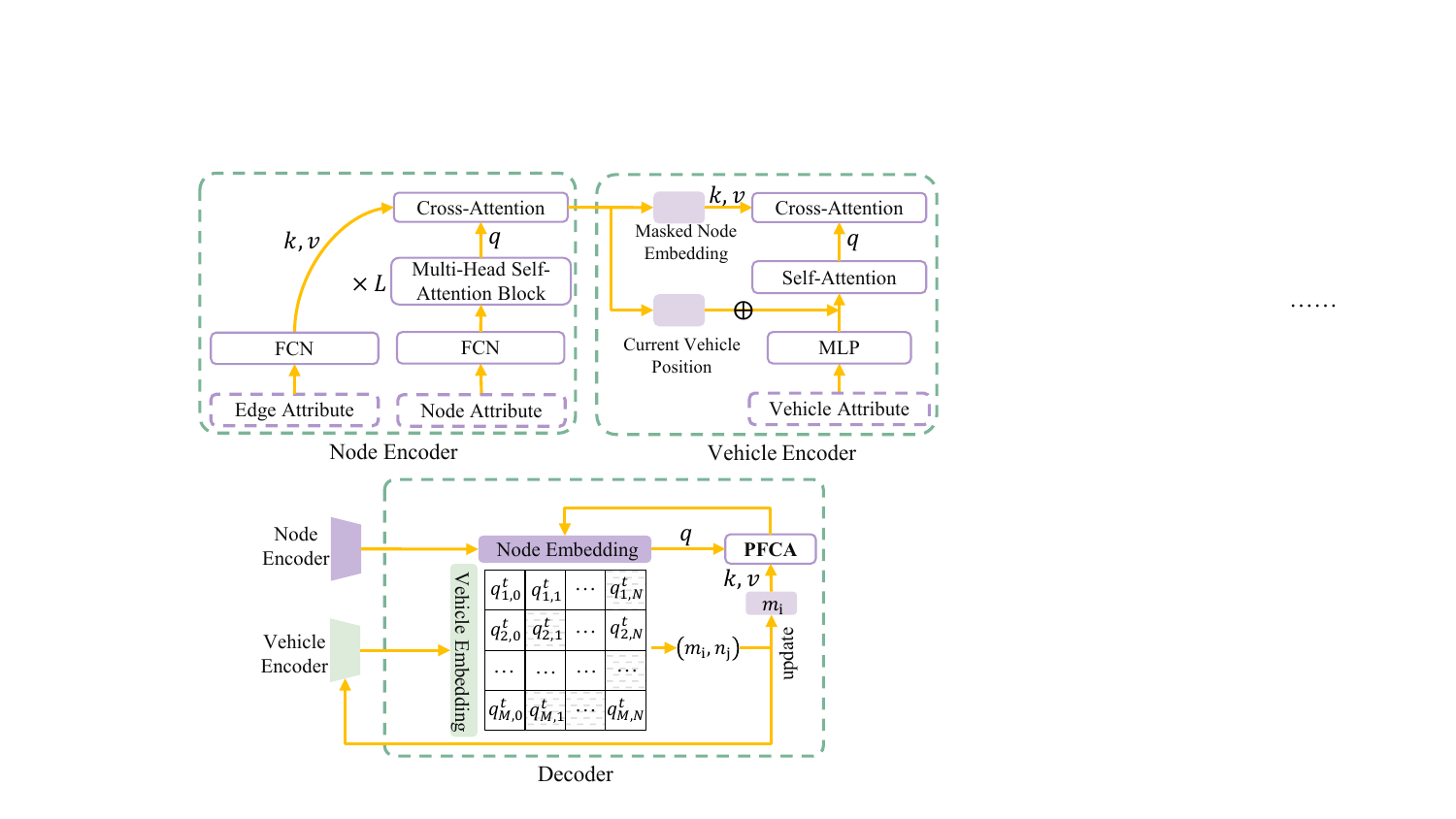}
	\caption{Overview of the proposed ECHO solver. To capture local topological information, ECHO  exploits the proposed dual-modality node encoder to fuse node and edge features. Subsequently, the vehicle encoder integrates the node embedding with vehicle-specific attributes to produce the vehicle embedding. Finally, to highlight the importance of the vehicle selected at the $t-1$th time step, ECHO employs a decoder incorporating the proposed PFCA mechanism to efficiently select vehicle-node pairs.} 
	\label{figecho}
\end{figure*}
The illustration of ECHO is schematically presented in Figure~\ref{figecho}. In this section, we first introduce the customized dual-modality node encoder and the adopted vehicle encoder for MMHCVRP, then detail the PFCA mechanism-based decoder, and finally present the proposed data augment method along with the corresponding RL training algorithm. The characteristics of different NCO solvers are summarized in Table~\ref{tablecom}. Compared to 2D-Ptr \citep{2dptr} and PARCO \citep{parco}, our ECHO achieves the new SOTA performance for NCO solvers (see the experimental results in Section~\ref{sec5}). This improvement can be primarily attributed to its architectural design, which explicitly leverages edge attributes and incorporates information from historically selected vehicles, whereas existing NCO solvers tailored for MMHCVRP do not account for these aspects. The source code of ECHO is available online\footnote{https://github.com/wuuu110/echo}.

\begin{table}[!t]
    \centering 
        \caption{Summary of NCO solvers tailored for MMHCVRP}
    \resizebox{\textwidth}{!}{
    \begin{tabular}{c|c |c |c|c |c} \toprule
      Method   & Type & Edge Attribute & Historical Vehicle & Data Augment & Performance\\ \midrule
      2D-Ptr  & AR &\ding{55}  & \ding{55}   & \ding{55} & large gap\\
      PARCO  & PAR & \ding{55} & \ding{55}  & node & large gap \\
      ECHO  & AR & \ding{51} & \ding{51}  & node \& vehicle & smallest gap\\ \toprule
    \end{tabular}}

    \label{tablecom}
\end{table}

\subsection{Node and Vehicle Encoders} 
Unlike TSP and CVRP, MMHCVRP more faithfully represents real-world routing scenarios by involving multiple vehicles. Accordingly, alongside the node encoder, ECHO incorporates a dedicated vehicle encoder to capture vehicle-specific attributes. In this subsection, we present the architectures of the node and vehicle encoders, respectively. 

\label{sec4.1}
\subsubsection{Node Encoder}

To capture the local topological relationships among nodes while mitigating interference from irrelevant nodes, we propose a dual-modality node encoder, which integrates edge attributes (i.e., topological distances) with node attributes (i.e., coordinates and demands). Specifically, ECHO first exploits two FCNs to embed the initial node and edge features, respectively. Let $u_{j}= (x_j,y_j,d_j^t)$ denote the initial attributes of the $j$th node. The node feature $U \in \mathbb{R}^{(N+1) \times d}$ is computed according to Eq.~(\ref{eq7}).
\begin{equation}
\label{eq7}
    U_{j} = 
\begin{cases} 
u_{j} \bm{W}_{no} + \text{DT}, & j=0, \\
u_{j} \bm{W}_{no}, & \text{otherwise.} 
\end{cases}
\end{equation}
where the trainable parameters include $\bm{W}_{no} \in \mathbb{R}^{3 \times d}$ and the depot token $\text{DT} \in \mathbb{R}^{d}$. Following the prior study \citep{2dptr}, we exploit a learnable depot token to distinguish the depot from customer nodes during training. 
{Let $e\in \mathbb{R}^{(N+1) \times (N+1)}$ denote the pairwise Euclidean distance matrix, which includes all customer nodes as well as the depot. In addition, no additional normalization is applied in our experiments, as the node coordinates are sampled from the unit square $[0,1]\times[0,1]$. To compute the edge features $E^K \in \mathbb{R}^{(N+1) \times d}$ and $E^V \in \mathbb{R}^{(N+1) \times d}$, the matrices $B^K,B^V \in \mathbb{R}^{(N+1) \times (N+1) \times d}$ are first computed from the pairwise distances according to Eq.~(\ref{eq8}). Subsequently, an attention-based aggregation over neighboring edges is performed to obtain $E^K$ and $E^V$, defined in Eqs.~(\ref{eq8+})$\sim$(\ref{eq8-}).

\begin{equation}\label{eq8}
    B^K= e\bm{W}^K_{ed}, \quad  B^V= e\bm{W}^V_{ed},
\end{equation}
\begin{equation}\label{eq8+}
E_i^K =
\sum_{j=0}^{N}
A^K_{ij} B^K_{ij},  \quad  E_i^V =
\sum_{j=0}^{N}
A^V_{ij} B^V_{ij} 
\end{equation}
\begin{equation}\label{eq80}
    A^K_{ij} =
\frac{\exp(C^K_{ij})}
{\sum_{k=0}^{N}\exp(C^K_{ik})}, \quad  A^V_{ij} =
\frac{\exp(C^V_{ij})}
{\sum_{k=0}^{N}\exp(C^V_{ik})}, 
\end{equation}
\begin{equation} \label{eq8-}
    C^K = B^K\bm{W}^K_{s},  \quad  C^V = B^V\bm{W}^V_{s}, 
\end{equation}
where $i, j \in \{0, \cdots, N\}$, and the trainable parameters include $\bm{W}^K_{ed},\bm{W}^V_{ed}$,$ \bm{W}^K_{s}$, and $\bm{W}^V_{s} \in \mathbb{R}^{1 \times d}$. The pseudocode of the edge feature computation process is presented in Algorithm~\ref{alg1}. Notably, this edge embedding mechanism does not introduce parameters that scale with the number of nodes $N$. Therefore, the same set of trainable parameters can be applied to instances of different sizes without padding or a fixed maximum $N$, enabling the model to naturally generalize across problem scales. In practice, the edge features are recomputed for each instance based on its coordinates, ensuring that the encoder captures instance-specific topological relationships.

\begin{algorithm}[!t]
\caption{Edge Feature Computation}
\label{alg1}
\begin{algorithmic}[1]

\STATE \textbf{Input:} Euclidean distance matrix $e$
\STATE \textbf{Output:} Edge-aware node features $E^K$ and $E^V$

\STATE $B^K \gets \text{Linear}(e.\text{unsqueeze}(-1))$ \quad $B^V \gets \text{Linear}(e.\text{unsqueeze}(-1))$ 
\STATE \textcolor{gray}{\# [batch, N+1, N+1, d] \hfill see Eq.~(\ref{eq8})}

\STATE $C^K \gets \text{Linear}(B^K).\text{squeeze}(-1)$ \quad $C^V \gets \text{Linear}(B^V).\text{squeeze}(-1)$
\STATE \textcolor{gray}{\# [batch, N+1, N+1] \hfill see Eq.~(\ref{eq8-})}

\STATE $A^K \gets \text{softmax}(C^K,\text{ dim}=2)$ \quad $A^V \gets \text{softmax}(C^V,\text{ dim}=2)$
\STATE \textcolor{gray}{\#  [batch, N+1, N+1] \hfill see Eq.~(\ref{eq80})}

\STATE $E^K \gets \text{sum}(B^K \cdot A^K.\text{unsqueeze}(-1),\text{ dim}=2)$
\STATE $E^V \gets \text{sum}(B^V \cdot A^V.\text{unsqueeze}(-1),\text{ dim}=2)$
\STATE \textcolor{gray}{\#  [batch, N+1, d] \hfill see Eq.~(\ref{eq8+})}

\STATE \textbf{return} $E^K, E^V$

\end{algorithmic}
\end{algorithm}
Then, the node encoder exploits the number of $L$ stacked Multi-Head Self-Attention Blocks to facilitate information exchange among node attributes. Let $H^0=\{U_{0}, \cdots, U_{N}\}$, the node embedding $H^{l+1}$ is computed according to Eq.~(\ref{eq9}).
\begin{equation}
\label{eq9}
\bar{H}^{l} = \text{BN}\left( H^{l} + \text{MHA}_{\text{self}}\left( H^{l}, H^{l}, H^{l} \right) \right),
\end{equation}
\begin{equation}
H^{l+1} = \text{BN}\left( \bar{H}^{l} + \text{FF}\left( \bar{H}^{l} \right) \right),
\end{equation}
where $\text{BN}$ denotes the batch normalization. The functions $\text{FF}$ and $\text{MHA}_{\text{self}}$  denote a two-layer feed‑forward network with ReLU activation function and the multi-head self-attention mechanism, respectively. They are defined in Eqs.~(\ref{eq11})$\sim$(\ref{eq12}).
\begin{equation}
\label{eq11}
\text{FF}(\bar{H}^{l})= \Bigl(\text{ReLU}\bigl( \bar{H}^{l}\bm{W}_1 \bigr) \Bigr) \bm{W}_2, 
\end{equation}
\begin{equation}
\label{mha}
    \text{MHA}_{\text{self}}(Q,K,V)=\left( ||_{k=1}^{8} \text{ATT}_{k}(Q,K,V) \right) \bm{W}^{O},
\end{equation}
\begin{equation}
\label{eq12}
    \text{ATT}_{k}(Q,K,V)=\operatorname{softmax}\left(\frac{\left(Q \bm{W}_{Q,k}\right) \left(K \bm{W}_{K,k}\right)^{T}}{\sqrt{d'}}\right) V\bm{W}_{V,k},
\end{equation}
where $\bm{W}_1 \in \mathbb{R}^{d \times 4d}$, $\bm{W}_2 \in \mathbb{R}^{4d \times d}$, $\bm{W}^{O} \in \mathbb{R}^{d \times d}$ and $\bm{W}_{Q}, \bm{W}_{K}, \bm{W}_{V}\in \mathbb{R}^{d \times d'}$ are trainable parameters, and the number of head is set to 8, i.e., $d'=d/8$. The symbol $||$ denotes the concatenation operator. 

Subsequently, the node encoder exploits the vanilla cross-attention mechanism to integrate the extracted local topological features $E^K \in \mathbb{R}^{(N+1) \times d}$ and $E^V \in \mathbb{R}^{(N+1) \times d}$ (computed according to Eqs.~(\ref{eq8})$\sim$(\ref{eq8-})) with the node embedding $H^{L}\in \mathbb{R}^{(N+1) \times d}$ produced by the multi‑head self‑attention blocks, defined in Eq.~(\ref{eq14}).
\begin{equation}
\label{eq14}
\mathbf{N}= H^{L} + \alpha \text{MHA}_{\text{cross}}\left( H^{L}, E^K, E^V \right) ,
\end{equation}
where $\text{MHA}_{\text{cross}}$ is computed analogously to $\text{MHA}_{\text{self}}$ in Eq.~(\ref{mha}), except that the query $Q$ is derived from $H^L$ while the key $K$ and value $V$ are derived from $E^K$ and  $E^V$. The symbol $\alpha$ denotes the gating coefficient, defined in Eq.~(\ref{eq15}).
\begin{equation}\label{eq15}
    \alpha = \text{Sigmoid}\Bigr( \bigr(\text{MHA}_{\text{cross}}(H^{L}, E^K, E^V)||H^L \bigr) \bm{W}_{g}\Bigr),
\end{equation}
where $\bm{W}_{g} \in \mathbb{R}^{2d \times 1}$. By adopting this design, ECHO efficiently integrates local topological features into the node embedding, hence achieves SOTA performance compared with solvers (e.g., 2D-Ptr \citep{2dptr} and PARCO \citep{parco}) that do not incorporate such features (see Tables~\ref{s4_tablehcvrp_1}$\sim$\ref{s4_tablehcvrp_3}). It is worth mentioning that while certain studies \citep{wang2024distance, reevo} also highlight the importance of local topological relationships, ECHO departs fundamentally in its methodology. Specifically, our ECHO learns to aggregate the local topological features in the embedding space, whereas these prior studies reshape attention scores among nodes during decoding using these features. 
 
\subsubsection{Vehicle Encoder} 


Because each vehicle's current location corresponds to a node’s coordinates, the vehicle embedding should inherently incorporate the associated node embedding. Therefore, the vehicle encoder must fuse these node embeddings with vehicle‑specific attributes to produce the vehicle embedding. To this end, we adopt the vehicle encoder architecture proposed by the prior study \citep{2dptr}. Specifically, similar to the node encoder, the vehicle encoder first embeds the vehicle attributes (i.e., $\chi_i, \rho_i, \hat{\rho}_i^t,$ and $\delta^t_i$, see Table~\ref{tabparams} for definitions about these attributes), using two fully connected layers with ReLU activations, defined in Eq.~(\ref{eq16}).
\begin{equation}\label{eq16}
        \mathbf{M}_{i}^{1,t} = 
    \Bigl[ \text{ReLU}\Bigl( \bigl( \chi_i, \rho_i, \hat{\rho}_i^t, \delta^t_i \bigr) \bm{W}_3 \Bigr) \Bigr] \bm{W}_4 
    + \text{PE}_{i}^{t} \bm{W}_{pe}
\end{equation}
where $\bm{W}_3 \in \mathbb{R}^{d \times 4d}$, $\bm{W}_4 \in \mathbb{R}^{4d \times d}$, and $\bm{W}_{pe} \in \mathbb{R}^{d \times d}$ are trainable parameters. $\text{PE}_{i}^{t}$ denotes the embedding of the node most recently visited by the $i$th vehicle at the $t$th time step. It is worth mentioning that, unlike the node encoder, the vehicle encoder needs to update vehicle embeddings throughout the decoding process to reflect changes in vehicle attributes.

Subsequently, the vehicle encoder applies a self‑attention layer to facilitate inter‑vehicle communication, followed by a vanilla cross‑attention layer to infuse each vehicle embedding with contextual information from node embeddings. The vehicle embedding $\mathbf{M}^{t}$ is computed according to Eqs.~(\ref{eq17}) and (\ref{eq18}).
\begin{equation}
\label{eq17}
    \mathbf{M}^{2,t} = \mathbf{M}^{1,t} + \text{MHA}_{\text{self}}\left( \mathbf{M}^{1,t}, \mathbf{M}^{1,t}, \mathbf{M}^{1,t} \right),
\end{equation}
\begin{equation}
\label{eq18}
    \mathbf{M}^{t} = \mathbf{M}^{2,t} + \text{MHA}_{\text{cross}}\left( \mathbf{M}^{2,t}, \mathbf{N}^{\text{mask}}, \mathbf{N}^{\text{mask}} \right),
\end{equation}
where $\text{MHA}_{\text{self}}$ and $\text{MHA}_{\text{cross}}$ denote the self-attention and cross-attention operations, respectively, $\mathbf{N}^{\text{mask}}$ denotes the node embedding with all visited nodes masked out. Finally, the vehicle embedding $\mathbf{M}^{t}$ and $\mathbf{N}$ are fed into the decoder to effectively select the vehicle-node pair at each time step.

\subsection{Decoder of ECHO}
\label{sec4.2}
To efficiently select vehicle-node pairs, we design a decoder inspired by the CLIP architecture \citep{clip}. While the prior study \citep{2dptr} also employed a similar CLIP-based architecture, our decoder incorporates fundamentally different structural components. In particular, we propose a \textbf{Parameter-Free Cross Attention (PFCA)} mechanism, which explicitly amplifies the influence of the node selected at the $t-1$th time step, thereby reducing myopic decision-making. To the best of our knowledge, ECHO is the first work to explicitly model historically selected vehicle information to alleviate myopic decisions.

\begin{figure}[!t]
	\centering
	\includegraphics[scale=0.65]{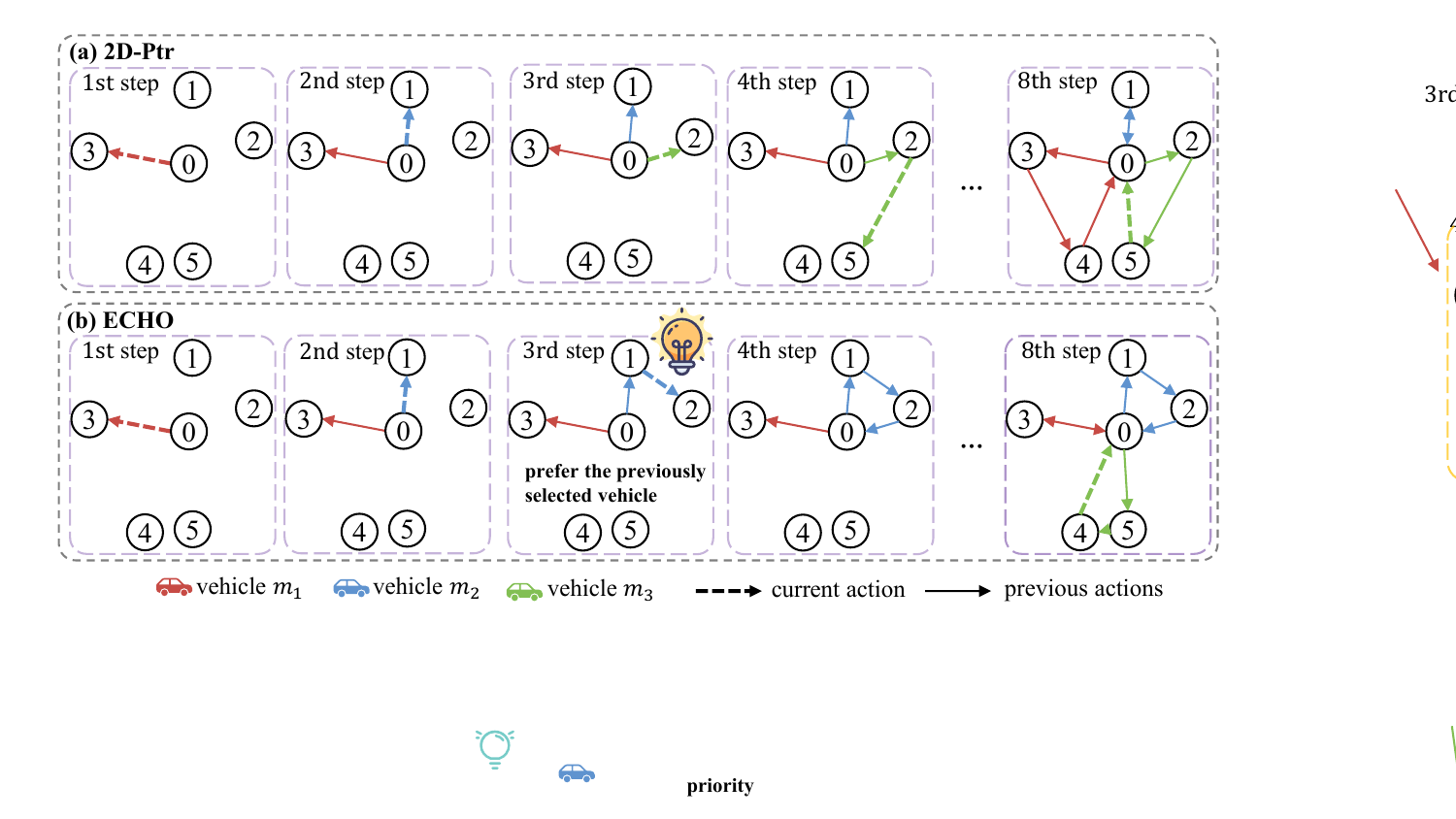}
	\caption{Illustration on the decoding processes of (a) 2D-Ptr and (b) ECHO. 2D-Ptr myopically relies on cumulative travel time to select the vehicle-node pair at each time step, which may lead to suboptimal solutions. Whereas the proposed ECHO alleviates this problem by emphasizing the vehicle selected in the preceding time step.} \label{fig_comparison}
\end{figure}

Specifically, given the vehicle embedding $\mathbf{M}^t\in\mathbb{R}^{M\times d}$ and node embedding $\mathbf{N}\in\mathbb{R}^{(N+1) \times d}$ produced by their respective encoders, the proposed decoder first integrates the vehicle embedding selected at the $t-1$th time step, $\mathbf{M}_s$, with the node embedding $\mathbf{N}$ to yield the updated node embedding $\hat{\mathbf{N}}^t$. For the initial step ($t \neq 0$),  no vehicle has been selected previously; therefore, $\mathbf{M}_s$ is undefined and $\hat{\mathbf{N}}^t$ is directly set to $\mathbf{N}$. For $t > 0$, $\hat{\mathbf{N}}^t$ is computed via the proposed PFCA mechanism, which injects the historical vehicle information into the node embeddings. Formally, $\hat{\mathbf{N}}^t$ is computed according to Eqs.~(\ref{eq19}) and (\ref{eq20}).
\begin{equation}
\label{eq19}
\hat{\mathbf{N}}^t=
\begin{cases}
\mathbf{N}, & t = 0, \\
\text{PFCA}(\mathbf{M}_s ,\mathbf{N}), & \text{otherwise},
\end{cases}
\end{equation} 
\begin{equation}
\label{eq20}
\text{PFCA}(\mathbf{M}_s ,\mathbf{N}) = \text{softmax}(\frac{\mathbf{N}  {(\mathbf{M}_s) }^\top}{\sqrt{d}}) \mathbf{M}_s + \mathbf{N}. 
\end{equation} 
 By applying the proposed PFCA mechanism, information from the previously selected vehicle is seamlessly integrated into the node embedding. The advantage of ECHO over 2D-Ptr is illustrated in Figure~\ref{fig_comparison}.


Subsequently, the vehicle embedding $\mathbf{M}^t$ and the updated node embedding $\hat{\mathbf{N}}^t$ are combined to compute the attention scores for all vehicle-node pairs at the current time step. In addition, to ensure feasibility, we mask invalid pairs by setting them to $-\infty$. For example, if a vehicle lacks sufficient remaining capacity to fulfill a node’s demand, its corresponding vehicle-node pair is masked. Formally, the attention score $\beta_{i,j}^t$ is defined according to Eq.~(\ref{eq21}).
\begin{equation}
\label{eq21}
\beta_{i,j}^t = 
\begin{cases}
-\infty, & d_j^t = 0 \land j \neq 0, \\
-\infty, &  \ell_i^t = 0 \land j = 0, \\
-\infty, &  \rho_i - \hat{\rho}_i^t < d_j^t, \\
\lambda \tanh\left( \dfrac{\mathbf{M}_i^t \hat{\mathbf{N}}_j^t}{\sqrt{d}} \right), & \text{otherwise},
\end{cases}
\end{equation}
where $\tanh(\cdot)$ denotes the hyperbolic tangent activation function, $\lambda$ denotes the hyper-parameter set identically to that in the prior study \citep{2dptr}.

Finally, the probability of selecting vehicle-node pair $(m_i,n_j)$ at the $t$th time step is defined in Eq.~(\ref{eq22}).
\begin{equation}
\label{eq22}
     p_{i,j}^t = \text{softmax}(\beta_{i,j}^t).
\end{equation}
After computing the probability distribution, ECHO constructs solutions for the given MMHCVRP instance using either Greedy or Sampling strategies (see Section~\ref{sec5.1} for more details on these strategies). Moreover, if the $i$th vehicle is selected at $t$th step, the stored previous-vehicle embedding is updated as $\mathbf{M}_s = \mathbf{M}^{t}_i$, which is then used to guide the decision in the succeeding time step. It is worth noting that the decoding strategy only affects how the vehicle–node pair is selected (i.e., argmax for greedy decoding or probabilistic sampling), whereas the update and propagation of $\mathbf{M}_s$  follow an identical procedure once the vehicle has been determined.

\subsection{Data Augment Method}
\label{sec4.3}
\begin{figure}[!t]
	\centering
\includegraphics[scale=1]{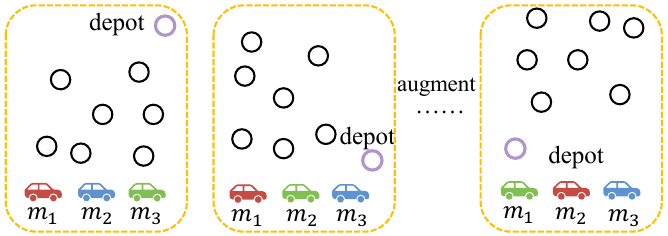}
	\caption{Illustration of the proposed augment method.} 
	\label{figaugment}
\end{figure}

Data augment techniques have been widely applied to stabilize the RL-based training process of NCO solvers in canonical single-vehicle VRPs \citep{pomo, sym-nco}. However, in the single‑vehicle setting, these methods need only exploit the node symmetry of the VRP. In contrast, MMHCVRP exhibits both node symmetry and vehicle‐permutation invariance. To the best of our knowledge, no prior work has simultaneously leveraged these two properties when solving the MMHCVRP, which may underlie their suboptimal performance.

\begin{table}[!t]
\centering
\caption{Node Augment Transformations}  
\label{tablenode}
\begin{tabular}{cc}
\toprule
\multicolumn{2}{c}{$f(x, y)$} \\  
\midrule
1: $(x, y)$       & 2: $(y, x)$       \\  %
3: $(x, 1 - y)$   & 4: $(y, 1 - x)$   \\
5: $(1 - x, y)$   & 6: $(1 - y, x)$   \\
7: $(1 - x, 1 - y)$ & 8: $(1 - y, 1 - x)$ \\
\bottomrule
\end{tabular}
\end{table}

As shown in Figure~\ref{figaugment}, to address this open challenge, we propose a tailored data augment method that integrates vehicle permutation invariance and node symmetry. Specifically, the augmentation is performed in two ordered steps: (1) node-level geometric reflection, followed by (2) vehicle permutation. Following the node augment approach of prior study  \citep{pomo}, we generate $K=8$ augmentations per instance by reflecting node coordinates across multiple axes, as shown in Table~\ref{tablenode}. During this step, only spatial coordinates are transformed, while the node demands remain unchanged and are directly replicated across the augmented instances to preserve problem feasibility.

After node reflection, we perform vehicle-level augmentation by sampling a random permutation over the vehicle indices. Each vehicle is assigned a unique identifier, and the permutation only reorders the vehicle indices while keeping their intrinsic attributes (e.g., capacities and speeds) unchanged, thereby explicitly exploiting the vehicle-permutation invariance. For example, in a case with three vehicles $m_1, m_2,$ and $m_3$, there exist $M!$ number of possible vehicle permutations, including $(m_1, m_2, m_3)$, $(m_1, m_3, m_2)$, $(m_2, m_1, m_3)$, $(m_2, m_3, m_1)$, $(m_3, m_2, m_1)$, and $(m_3, m_1, m_2)$. Notably, when $M!$ is smaller than $K$, duplicate vehicle permutations across augmented samples are inevitable and thus allowed. The pseudocode of the overall data augment procedure is presented in Algorithm~\ref{alg2}. 

\begin{algorithm}[!t]
\caption{Data Augment  Procedure}
\label{alg2}
\begin{algorithmic}[1]
\STATE \textbf{Input:} Instance $G$, augmentation factor $K=8$,
\STATE \textbf{Output:} Augmented Instance Set  $\{G_{i}\}_{k=1}^{K}$
\FOR{$k=1$ to $K$}
\STATE Apply transformation to customers and depot coordinates according to Table~\ref{tablenode}.
\textcolor{gray}{\hfill\# The demands of Customers  keep unchanged} 
\STATE Sample a random permutation over the vehicle indices and reorder them accordingly.
 \ENDFOR
\STATE \textbf{return  $\{G_{i}\}_{k=1}^{K}$}  
 \end{algorithmic}
\end{algorithm}


We train ECHO using the well-known REINFORCE algorithm \citep{williams_simple_1992} with a shared baseline similar to the prior studies \citep{pomo, sym-nco}. As detailed in Section~\ref{sec3.1}, the objective of MMHCVRP is to minimize the maximum vehicle travel time. Accordingly, we define the total reward $R$ as the negative of this objective. The policy gradient is computed according to Eq.~(\ref{eq23}).
\begin{equation}
    \label{eq23}
        \nabla_\theta \mathcal{L} = \frac{1}{B \cdot K} \sum_{i=1}^{B} \sum_{j=1}^{K} 
        \Big(R({\tau}_{i,j}|{S}_i)-b({S}_i)\Big) \cdot \nabla_\theta \log p_\theta({\tau}_{i,j}|{S}_i),
\end{equation}

\noindent where $B$ and $K$ denote the batch size and the number of augmentations per instance, respectively, and $b(\cdot)$ denotes the shared baseline, computed as the average reward across all augmentations.

\section{Experimental Results}
\label{sec5}
To assess the performance of ECHO, we conduct extensive experiments in this section. Section~\ref{sec5.1} first describes datasets, baseline methods, and evaluation metrics. Section~\ref{sec5.2} then compares the performance of different NCO solvers. Next, Section~\ref{sec5.3} evaluates the generalization ability of ECHO across different scales and distribution patterns. Finally, Section~\ref{sec5.5} conducts ablation studies to assess the effectiveness of all proposed components.
\subsection{Experimental Settings}
\label{sec5.1}
\subsubsection{Datasets and Hardware configuration} Following the prior studies \citep{2dptr,drl}, we assess the performance of ECHO on MMHCVRP instances with varying numbers of vehicles and nodes. Except for the cross-distribution generalization experiments in Section~\ref{sec5.3}, the coordinates of customer nodes and depot node are uniformly sampled from the unit square $[0,1] \times [0,1]$, and the demand of customer nodes are uniformly sampled from the set $\{ 1, 2, \cdots, 9\}$. For vehicles, capacities are uniformly sampled from $\{20, 21, \cdots, 40\}$, and speeds are uniformly sampled from the interval $[0.5, 1]$. For each vehicle and node scale, the training dataset comprises the number of 1,280,000 instances, and the test dataset comprises 10,000 instances. All experiments are conducted on a computer equipped with an
INTEL(R) XEON(R) GOLD 6338 CPU and two NVIDIA RTX 4090 GPUs (24GB).  During inference, we employ only a single GPU.

\subsubsection{Hyper-parameter}
\begin{table}[!t]
\centering
\caption{Hyper-parameter settings}
\label{tab:hyperparams}
\begin{tabular}{lc}
\toprule
\textbf{Hyperparameter} & \textbf{Value} \\
\hline
Training epochs & 50 \\
Batches per epoch & 20{,}000 \\
Batch size & 64 \\
Optimizer & Adam \\
Initial learning rate & $0.0001$ \\
Learning rate decay & 0.995 \\
Number of attention heads & 8 \\
Embedding dimension $d$ & 128 \\
Number of encoder layers $L$ & 3 \\
Temperature parameter $\lambda$ & 10 \\
\toprule
\end{tabular}
\end{table}
To ensure a fair comparison, we follow the hyperparameter settings of the prior study \citep{2dptr} for ECHO. Specifically, the model is trained for 50 epochs, with each epoch comprising 20,000 batches, and each batch containing 64 randomly generated problem instances. We employ the Adam optimizer with an initial learning rate of 0.0001 and a decay factor of 0.995 per epoch.
Furthermore, gradient clipping is applied to stabilize the training process, with the maximum L2 norm set to 3.0. The number of heads in
all multi-head attention is set to 8, while the embedding dimension $d$ and the number $L$ of layers in the node encoder are set to 128 and 3,
respectively. The temperature parameter $\lambda$ is set to 10.

\begin{table*}[!t]
\caption{Performance Comparison of different Algorithms on HCVRP Instances across Varying Node Scales $N=60,100$ with a Fixed Vehicle Number $M=3$ }
\label{s4_tablehcvrp_1}
\resizebox{\textwidth}{!}{
\begin{tabular}{l| c c c | c c c }
\toprule
\multicolumn{1}{c|}{$M$} & \multicolumn{6}{c}{3} \\
\midrule
\multicolumn{1}{c|}{$N$} & \multicolumn{3}{c|}{60} & \multicolumn{3}{c}{100}  \\
\cmidrule[0.5pt](lr){1-1} \cmidrule[0.5pt](lr){2-4} \cmidrule[0.5pt](lr){5-7} 
\multicolumn{1}{c|}{Metric} & Obj. $\downarrow$ & Gap (\%) $\downarrow$ & Time (s) $\downarrow$ & Obj. $\downarrow$ & Gap (\%) $\downarrow$ & Time (s) $\downarrow$  \\
\midrule
SISR & 6.57 & - & 271 & 10.29 & - & 615  \\
GA    & 9.21 & 40.18 & 233 & 15.33 & 48.98 & 479 \\
SA    & 7.04 & 7.15  & 130 & 11.13 & 8.16  & 434  \\
\midrule
AM (ICLR'19)  (\textit{g.}) 	& 8.49 & 29.22 & 0.08 & 12.68 & 23.23 & 0.14  \\
ET (AAAI'24)  (\textit{g.}) 	& 7.58 & 15.37 & 0.15 & 11.74 & 14.09 & 0.25  \\
DPN (ICML'24) (\textit{g.})	& 7.50 & 14.16 & 0.18 & 11.54 & 12.15 & 0.30  \\
DRL (TCYB'22)  (\textit{g.})	& 7.43 & 13.09 & 0.19 & 11.44 & 11.18 & 0.32   \\
2D-Ptr (AAMAS'24) (\textit{g.})          & 7.20 & 9.59 & 0.11 & 11.12 & 8.07 & 0.18  \\
PARCO (NeurIPS'25) (\textit{g.}) &7.12 & 8.37 & 0.04 & 10.98 & 6.71 & 0.06   \\
ECHO (ours) (\textit{g.}) 				& 6.82 & 3.94 & 0.12  & 10.68 & 3.82  & 0.20 \\
\midrule
AM (ICLR'19)  (\textit{s.}) 	& 7.62 & 15.98 & 0.14 & 11.82 & 14.87 & 0.29  \\
ET (AAAI'24) (\textit{s.}) 	& 7.14 & 8.68 & 0.21 & 11.20 & 8.84 & 0.41   \\
DPN (ICML'24)  (\textit{s.}) 	& 7.08 & 7.76 & 0.25 & 11.04 & 7.29  & 0.48 \\
DRL (TCYB'22)  (\textit{s.}) 	& 6.97 & 6.09 & 0.30 & 10.90 & 5.93  & 0.60   \\
2D-Ptr (AAMAS'24) (\textit{s.})  	& 6.82 & 3.81 & 0.13 & 10.71 & 4.08  & 0.22  \\
PARCO (NeurIPS'25)  (\textit{s.}) & 6.82 & 3.81 & 0.05 & 10.61 & 3.11 & 0.08  \\
ECHO (ours) (\textit{s.}) 		& \textbf{6.63} & \textbf{1.00} & 0.40   & \textbf{10.44} & \textbf{1.48} & 0.96  \\
\bottomrule
\end{tabular}}
\end{table*}
\begin{table*}[!h]
\caption{Performance Comparison of different Algorithms on HCVRP Instances across Varying Node Scales $N=60,100$ with a Fixed Vehicle Number $M=5$ }
\label{s4_tablehcvrp_2}
\resizebox{\textwidth}{!}{
\begin{tabular}{l| c c c | c c c }
\toprule
\multicolumn{1}{c|}{$M$} & \multicolumn{6}{c}{5} \\
\midrule
\multicolumn{1}{c|}{$N$} & \multicolumn{3}{c|}{60} & \multicolumn{3}{c}{100}  \\
\cmidrule[0.5pt](lr){1-1} \cmidrule[0.5pt](lr){2-4} \cmidrule[0.5pt](lr){5-7} 
\multicolumn{1}{c|}{Metric} & Obj. $\downarrow$ & Gap (\%) $\downarrow$ & Time (s) $\downarrow$ & Obj. $\downarrow$ & Gap (\%) $\downarrow$ & Time (s) $\downarrow$  \\
\midrule
 SISR  & 4.00 & - & 274 & 6.17 & - & 623  \\
GA    & 6.89 & 72.25 & 320 & 10.93 & 77.15 & 623 \\
SA     & 4.39 & 9.75  & 289 &  6.80 & 10.21 & 557  \\
\midrule
AM (ICLR'19) (\textit{g.}) 	& 5.51 & 37.75 & 0.08 & 8.10 & 31.28 & 0.13 \\
ET (AAAI'24)  (\textit{g.})		& 4.76 & 19.00 & 0.17 & 7.25 & 17.50 & 0.25  \\
DPN (ICML'24) (\textit{g.})   	& 4.60 & 15.00 & 0.19 & 6.94 & 12.48 & 0.40  \\
DRL (TCYB'22) (\textit{g.}) 	& 4.71 & 17.75 & 0.22 & 7.06 & 14.42 & 0.37 \\
2D-Ptr (AAMAS'24) (\textit{g.}) & 4.48 & 12.00 & 0.11 & 6.75 & 9.40  & 0.18 \\
PARCO (NeurIPS'25)  (\textit{g.}) & 4.40 & 10.00 & 0.05 &6.61 & 7.13 & 0.05  \\
ECHO (ours) (\textit{g.})		& 4.20 & 5.08  &  0.12 &6.40 & 3.88  &  0.19  \\
\midrule
AM (ICLR'19)  (\textit{s.}) 	& 4.82 & 20.50 & 0.13 & 7.45 & 20.75 & 0.28 \\
ET (AAAI'24)  (\textit{s.}) 	& 4.46 & 11.50 & 0.22 & 6.85 & 11.02 & 0.38 \\
DPN (ICML'24)  (\textit{s.}) 	& 4.35 & 8.75  & 0.28 & 6.66 & 7.94  & 0.52  \\
DRL (TCYB'22) (\textit{s.}) 	& 4.34 & 8.50 & 0.36 & 6.65 & 7.78  & 0.76  \\
2D-Ptr (AAMAS'24) (\textit{s.})  		& 4.20 & 5.00  & 0.13 & 6.46 & 4.70  & 0.23  \\
PARCO (NeurIPS'25)  (\textit{s.}) & 4.17 & 4.25 & 0.05 &6.36 & 3.08 & 0.08  \\
ECHO (ours) (\textit{s.}) 				& \textbf{4.05} & \textbf{1.32} & 0.52 & \textbf{6.24} & \textbf{1.29} &   1.18 \\
\bottomrule
\end{tabular}}
\end{table*}
\begin{table*}[!h]
\caption{Performance Comparison of different Algorithms on HCVRP Instances across Varying Node Scales $N=60,100$ with a Fixed Vehicle Number $M=7$ }
\label{s4_tablehcvrp_3}
\resizebox{\textwidth}{!}{
\begin{tabular}{l| c c c | c c c }
\toprule
\multicolumn{1}{c|}{$M$} & \multicolumn{6}{c}{7} \\
\midrule
\multicolumn{1}{c|}{$N$} & \multicolumn{3}{c|}{60} & \multicolumn{3}{c}{100}  \\
\cmidrule[0.5pt](lr){1-1} \cmidrule[0.5pt](lr){2-4} \cmidrule[0.5pt](lr){5-7} 
\multicolumn{1}{c|}{Metric} & Obj. $\downarrow$ & Gap (\%) $\downarrow$ & Time (s) $\downarrow$ & Obj. $\downarrow$ & Gap (\%) $\downarrow$ & Time (s) $\downarrow$  \\
\midrule
 SISR  & 2.91 & - & 276 & 4.45 & - & 625  \\
GA    & 5.98 & 105.50 & 405 & 9.10 & 104.49 & 772 \\
SA    & 3.30 & 13.40  & 362 & 5.01 & 12.58  & 678  \\
\midrule

AM (ICLR'19)  (\textit{g.}) 	& 4.15 & 42.61 & 0.09 & 6.13 & 37.75 & 0.13  \\
ET (AAAI'24)  (\textit{g.}) 	& 3.58 & 23.02 & 0.16 & 5.23 & 17.53 & 0.26  \\
DPN (ICML'24)  (\textit{g.})	& 3.45 & 18.56 & 0.26 & 4.98 & 11.91 & 0.43  \\
DRL (TCYB'22) (\textit{g.})	& 3.60 & 23.71 & 0.25 & 5.38 & 20.90 & 0.43   \\
2D-Ptr (AAMAS'24) (\textit{g.})          & 3.31 & 13.75 & 0.11 & 4.92 & 10.56 & 0.17 \\
PARCO (NeurIPS'25)  (\textit{g.}) &3.25  &11.68  & 0.05 & 4.79 & 7.64 & 0.05   \\
ECHO (ours) (\textit{g.}) 				& 3.04 & 4.78 & 0.12  & 4.61 & 3.71 & 0.19 \\
\midrule
AM (ICLR'19)  (\textit{s.}) 	& 3.63 & 24.74 & 0.14 & 5.58 & 25.39 & 0.28  \\
ET (AAAI'24) (\textit{s.}) 	& 3.33 & 14.43 & 0.22 & 4.98 & 11.91 & 0.40   \\
DPN (ICML'24) (\textit{s.}) 	& 3.20 & 9.97 & 0.38 & 4.79 & 7.64  & 0.78 \\
DRL (TCYB'22)  (\textit{s.}) 	& 3.25 & 11.68 &0.43 & 4.98 & 11.91  & 0.92  \\
2D-Ptr (AAMAS'24) (\textit{s.})  		& 3.09 & 6.19 & 0.14 & 4.68 & 5.17  & 0.24  \\
PARCO (NeurIPS'25)  (\textit{s.}) & 3.06 & 5.15 & 0.07 & 4.58 & 2.92 & 0.09  \\
ECHO (ours) (\textit{s.}) 				& \textbf{2.93} & \textbf{1.02} & 0.63  & \textbf{4.49} & \textbf{0.91} & 1.45  \\
\bottomrule
\end{tabular}}
\end{table*}

\subsubsection{Baselines and Evaluation Metrics:} We compare ECHO against six NCO benchmarks, including AM \citep{kool_attention_2019}, DRL \citep{drl}, ET \citep{et}, DPN \citep{dpn}, 2D-Ptr \citep{2dptr}, and PARCO \citep{parco}. Of these, DRL, 2D-Ptr, and PARCO are specifically designed for the MMHCVRP, ET and DPN target the Min-max TSP, and AM is tailored to TSP and CVRP. In addition, following the prior studies \citep{2dptr,drl}, we also include the heuristic algorithms GA \citep{KARAKATIC2015519}, SA \citep{ILHAN2021100911}, and the SOTA SISR \citep{sisr} for comparison. The evaluation metrics include the average objective function values (Obj.), Gap, and inference time (Time) per problem instance, where Gap is defined as the relative difference in the "objective value" between each method and the SISR heuristic. In addition, for NCO solvers, we report the results under two decoding strategies, namely Greedy and Sampling decoding strategies. With Greedy decoding, the solver always selects vehicle(s) and node(s) with the maximum probability at each time step. Whereas the sampling strategy randomly selects vehicle(s) and node(s) utilizing the learned probability, constructs the number of $k$ candidate solutions, and ultimately reports the best among them. In all experiments conducted in the paper, $k$ is set to 1280, following the precedent established by \citep{kool_attention_2019}.

\subsection{Performance Comparison}
\label{sec5.2}
In Tables~\ref{s4_tablehcvrp_1}$\sim$\ref{s4_tablehcvrp_3}, we compare the performance of ECHO against various heuristics and NCO solvers on MMHCVRP instances with varying numbers of vehicles ($M=3,5,7$) and nodes ($N=60,100$). As shown in Tables~\ref{s4_tablehcvrp_1}$\sim$\ref{s4_tablehcvrp_3}, ECHO achieves SOTA performance across varying numbers of vehicles and nodes compared with the other NCO solvers. When adopting the sampling decoding strategy, ECHO reduces the average gap to 3.65\% and 2.54\% relative to the AR-based 2D-Ptr and PAR-based PARCO solvers, respectively. The high-level performance reflects the combined contributions of all proposed components and methods (see Section~\ref{sec5.5}).  Notably, even under greedy decoding, ECHO matches PARCO’s performance in gap achieved with sampling. Compared to the SOTA heuristic SISR, ECHO exhibits an average gap of approximately 1\%, while running over 100× faster, making it more suitable for time-sensitive real-world scenarios.
To assess the robustness of ECHO, we also conduct experiments using three different random seeds. The experimental results show that ECHO achieves an average Obj. of 6.63 with a standard deviation of 0.0001, demonstrating its robustness.

\subsection{Generalization Studies}
\label{sec5.3}
\begin{figure}[!t]
	\centering
\includegraphics[scale=0.35]{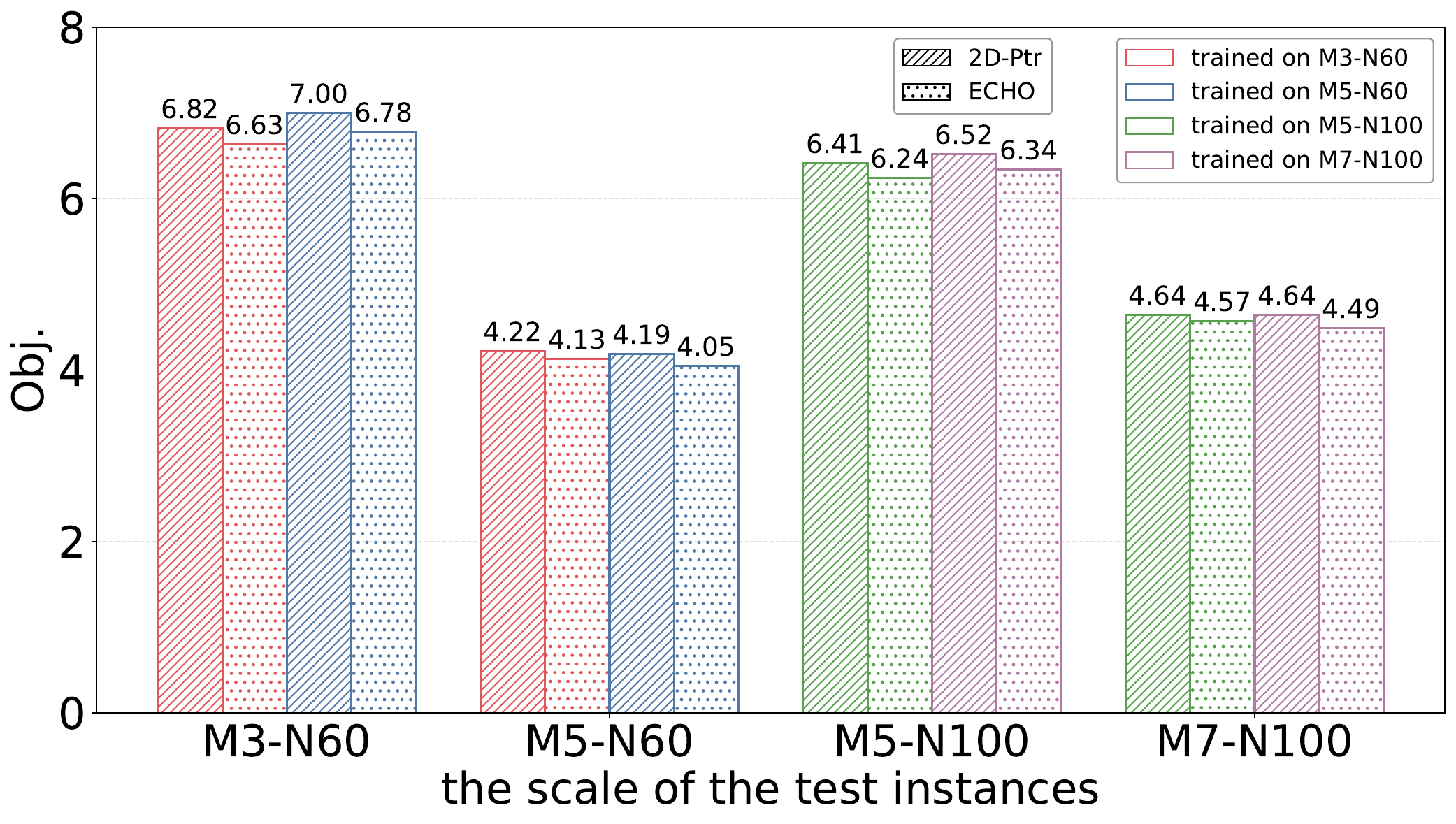}
	\caption{Generalization performance of NCO solvers across different vehicle scales.} 
	\label{vehicle}
\end{figure}

\begin{figure}[!t]
	\centering
\includegraphics[scale=0.4]{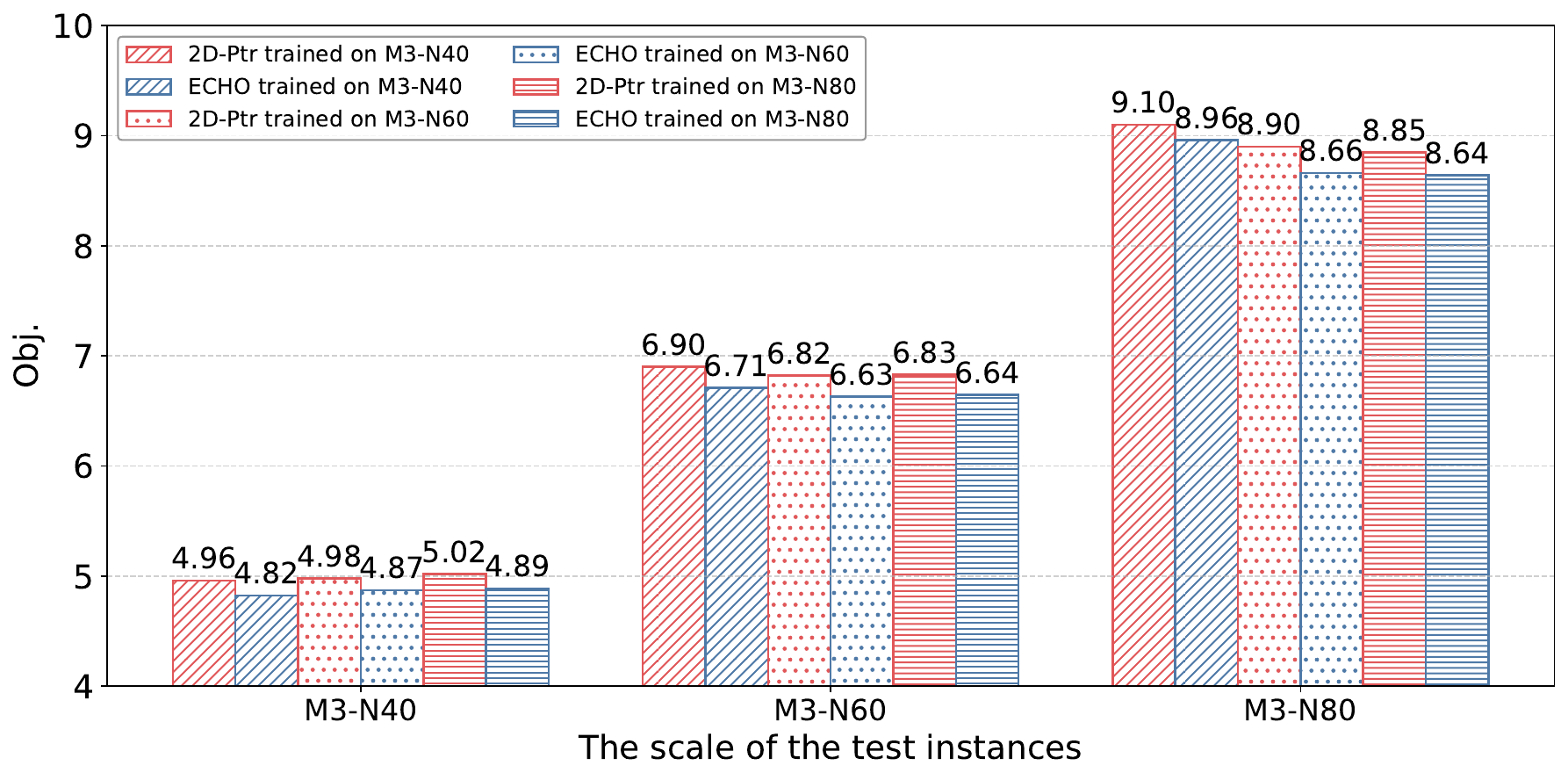}
	\caption{Generalization performance of NCO solvers across different node scales.} 
	\label{node}
\end{figure}
To assess the cross-scale and cross-distribution generalization abilities of ECHO, we conduct extensive experiments in this subsection. Specifically, in Figures.~\ref{vehicle} and \ref{node}, we present the generalization performance of ECHO versus 2D-Ptr across varying node and vehicle scales, respectively. The abbreviation M3-N60 denotes a configuration with three vehicles and 60 nodes. Furthermore, in Table~\ref{out_range}, we evaluate the generalization performance of ECHO and 2D-Ptr under three settings. Both ECHO and 2D-Ptr are trained on M7-N100 and then directly tested on these settings. Specifically, in Setting\_1 and Setting\_2,  we evaluate the solvers on instances with nine vehicles and 120 customer nodes, as well as seven vehicles and 100 customer nodes, respectively, which lie outside the range of the training configurations. To introduce more extreme heterogeneity regimes, in Settting\_3, the demands of customer nodes are sampled from a Gaussian distribution $\mathcal{N}(5, 4)$ and then discretized to integers within the set  $\{1,2,\cdots,9\}$, while vehicle capacities are uniformly sampled from  $\{20,21,\cdots,80\}$. Finally, in Figures~\ref{distribution-cluster} and ~\ref{distribution-explosion}, we assess the  generalization performance of ECHO and 2D-Ptr under  Clustered and Explosion distribution patterns, respectively. To generate Clustered test instances, following the procedure of the prior study \citep{invit}, we sample three cluster centers for each instance,  assign each node uniformly to one center, and perturb each node's coordinates with Gaussian noise to induce the clustered distribution. To generate Explosion test instances, following the procedure in the prior study \citep{invit}, we first generate node coordinates from a uniform distribution. Then, a disc is defined by uniformly selecting a center on the board and a radius sampled uniformly from $[0.1, 0.5]$. All nodes located inside this disc are subsequently displaced outward according to an exponential distribution with rate 10.

As shown in Figure~\ref{vehicle}, ECHO achieves comparable cross-vehicle scale generalization performance to 2D-Ptr. Furthermore, even when trained exclusively on non‑target vehicle scales, ECHO still outperforms 2D‑Ptr trained directly on the target scale. For example, when trained on the M5‑N60 configuration, ECHO attains an objective value of 6.78 on M3‑N60, compared to 6.82 achieved by 2D‑Ptr trained and evaluated on M3‑N60.
As shown in Figure~\ref{node}, ECHO also achieves high-level cross-node scale generalization ability compared to 2D-Ptr. As shown in Table~\ref{out_range}, ECHO outperforms 2D-Ptr across all settings, even when evaluated on instances whose scales lie outside the training configurations and under more extreme heterogeneity regimes.

\begin{table}[!t]
\footnotesize
    \centering
    \caption{Generalization performance of NCO solvers under Challenging Settings}
    \label{out_range}
    \begin{tabular}{l|c c c}
        \toprule
        \multirow{2}{*}{Solver} & \multicolumn{3}{c}{Obj. $\downarrow$} \\ \cmidrule[0.5pt](lr){2-4} 
          &  Setting\_1   & Setting\_2 & Setting\_3\\ \midrule
        2D-Ptr (AAMAS'24) & 4.27 & 6.80   & 3.54\\
       ECHO (ours) & \textbf{4.17}  & \textbf{6.57} & \textbf{3.51} \\\toprule
    \end{tabular}
\end{table}
\begin{figure}[!t]
	\centering
\includegraphics[scale=0.35]{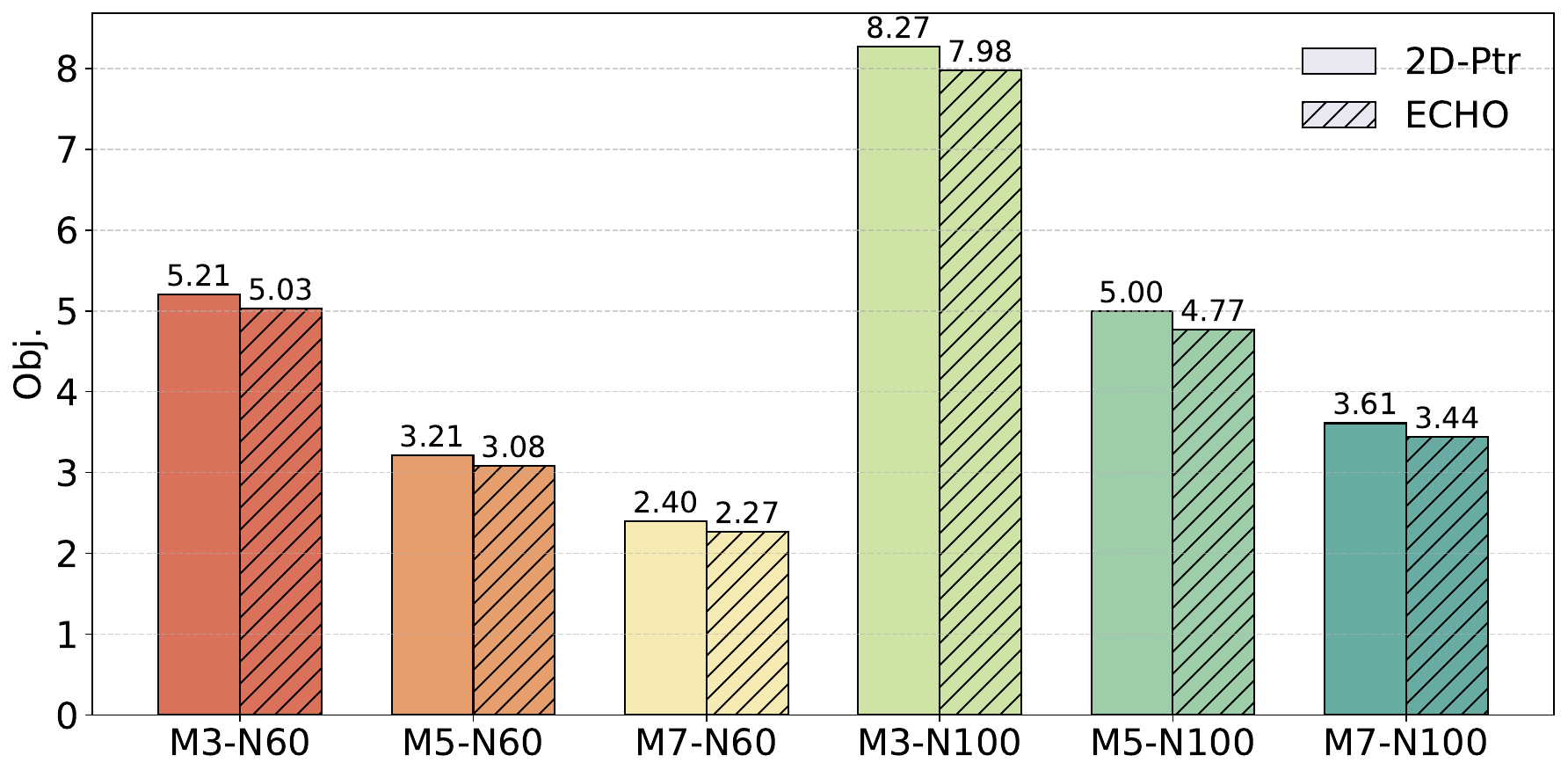}
	\caption{Generalization performance of NCO solvers under the Clustered distribution.} 
	\label{distribution-cluster}
\end{figure}
\begin{figure}[!t]
	\centering
\includegraphics[scale=0.35]{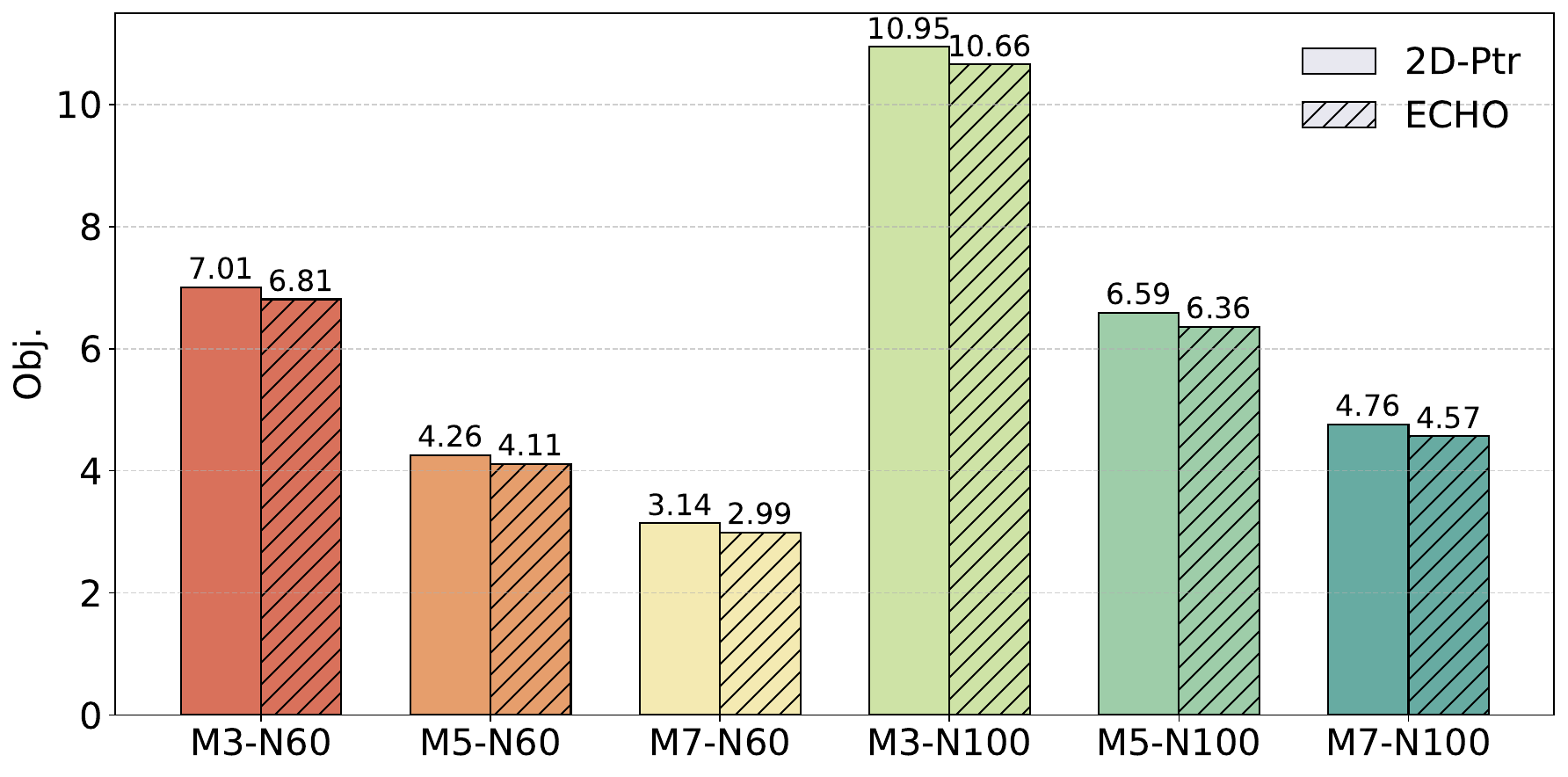}
	\caption{Generalization performance of NCO solvers under the Explosion distribution.} 
	\label{distribution-explosion}
\end{figure}

As shown in Figures~\ref{distribution-cluster} and \ref{distribution-explosion}, under the Clustered and Explosion distribution patterns, ECHO consistently outperforms 2D‑Ptr across all vehicle and node scales. This advantage arises from our dual‑modality encoder, which more effectively captures local topological relationships among nodes, thereby reducing sensitivity to distribution shifts.

\subsection{Ablation Studies}
\label{sec5.5}
In this subsection, we conduct ablation studies to investigate the contributions of key ECHO design components. The performance is evaluated on MMHCVRP instances with the number of three vehicles and 60 nodes, and the results are presented in Table~\ref{ablation}. Specifically, in w/o dual-modality encoder, the node encoder processes only node attributes, ignores edge attributes, and disables the cross‑attention operator used to fuse edge and node features. In w/o PFCA mechanism, the decoder omits historical vehicle information by disabling the PFCA mechanism. In w/o vehicle augment, only node data are augmented following the prior study \citep{pomo}, while vehicle data remain unchanged. As shown in Table~\ref{ablation}, the experimental results demonstrate that each component contributes to the SOTA performance of ECHO.
\begin{table}[!t]
    \footnotesize
    \centering
    \caption{Ablation Study Results on Different Design Choices}
    \label{ablation}
    \begin{tabular}{l|c}
        \toprule
        Algorithm & Gap (\%)   \\ \midrule
        w/o dual-modality encoder&  1.82 \\
        w/o PFCA mechanism  &  1.18\\ 
        w/o vehicle augment & 1.12\\ 
       ECHO &  1.00    \\\toprule
    \end{tabular}
\end{table} 

\subsection{Case Study}
To illustrate the effectiveness of the proposed PFCA mechanism, we visualize the visiting trajectories produced by 2D-Ptr and ECHO on the same instance. As shown in Figure~\ref{vis}, ECHO tends to extend the route of the vehicle selected in the previous step, which can be attributed to the PFCA mechanism. In contrast, 2D-Ptr occasionally ignores the priority of the vehicle selected in the preceding step, leading to worse performance (Obj. = 5.04). For instance, after assigning the blue vehicle to visit the 31st customer at the 42nd step, 2D-Ptr does not select this vehicle again until the 57th step.
\begin{figure}[!t]
\centering
\begin{subfigure}[b]{0.9\linewidth}
\centering
\includegraphics[width=\linewidth]{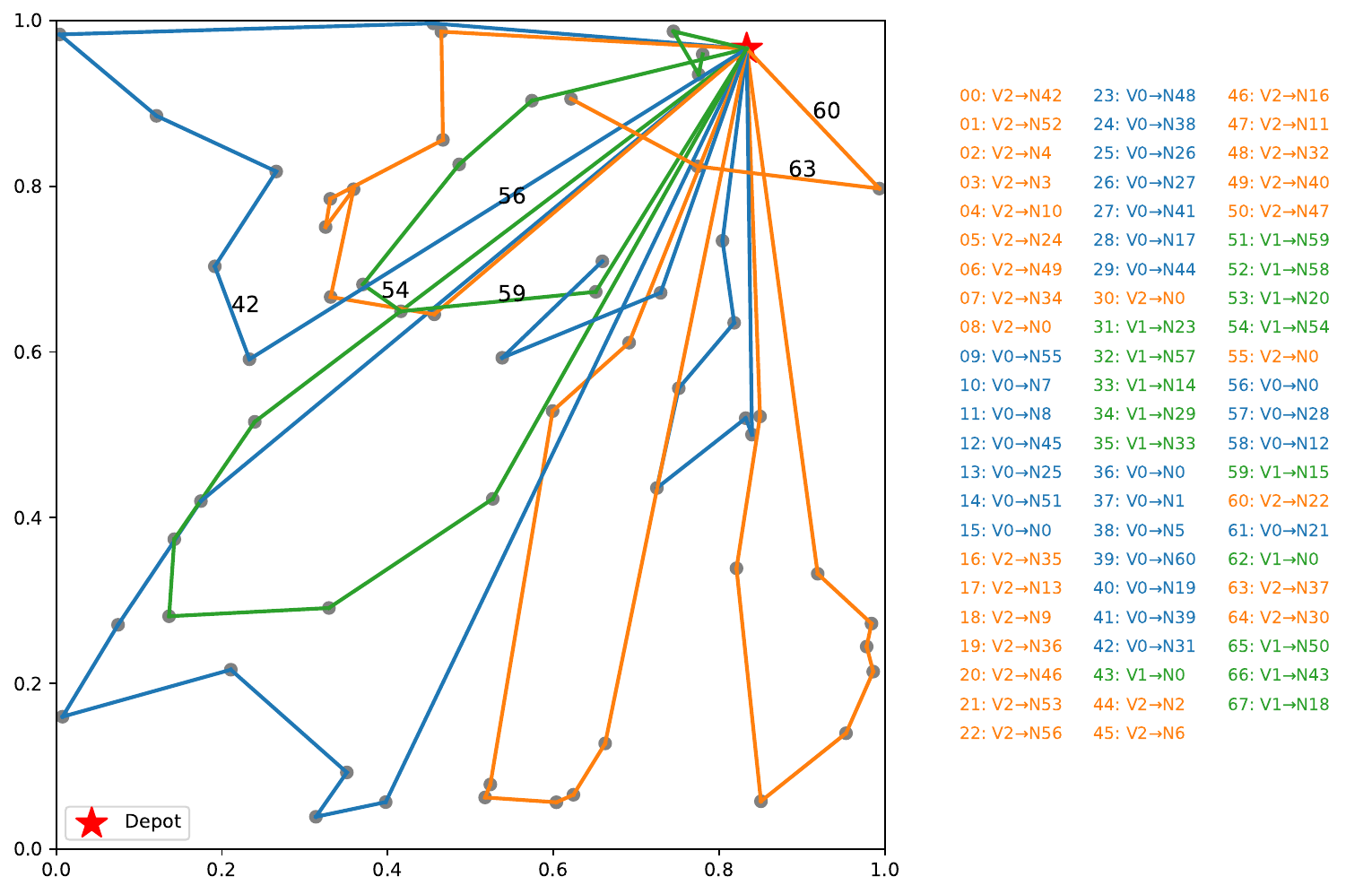}
\caption{2D-Ptr (Obj. 5.04)}
\end{subfigure}
\begin{subfigure}[b]{0.9\linewidth}
\centering
\includegraphics[width=\linewidth]{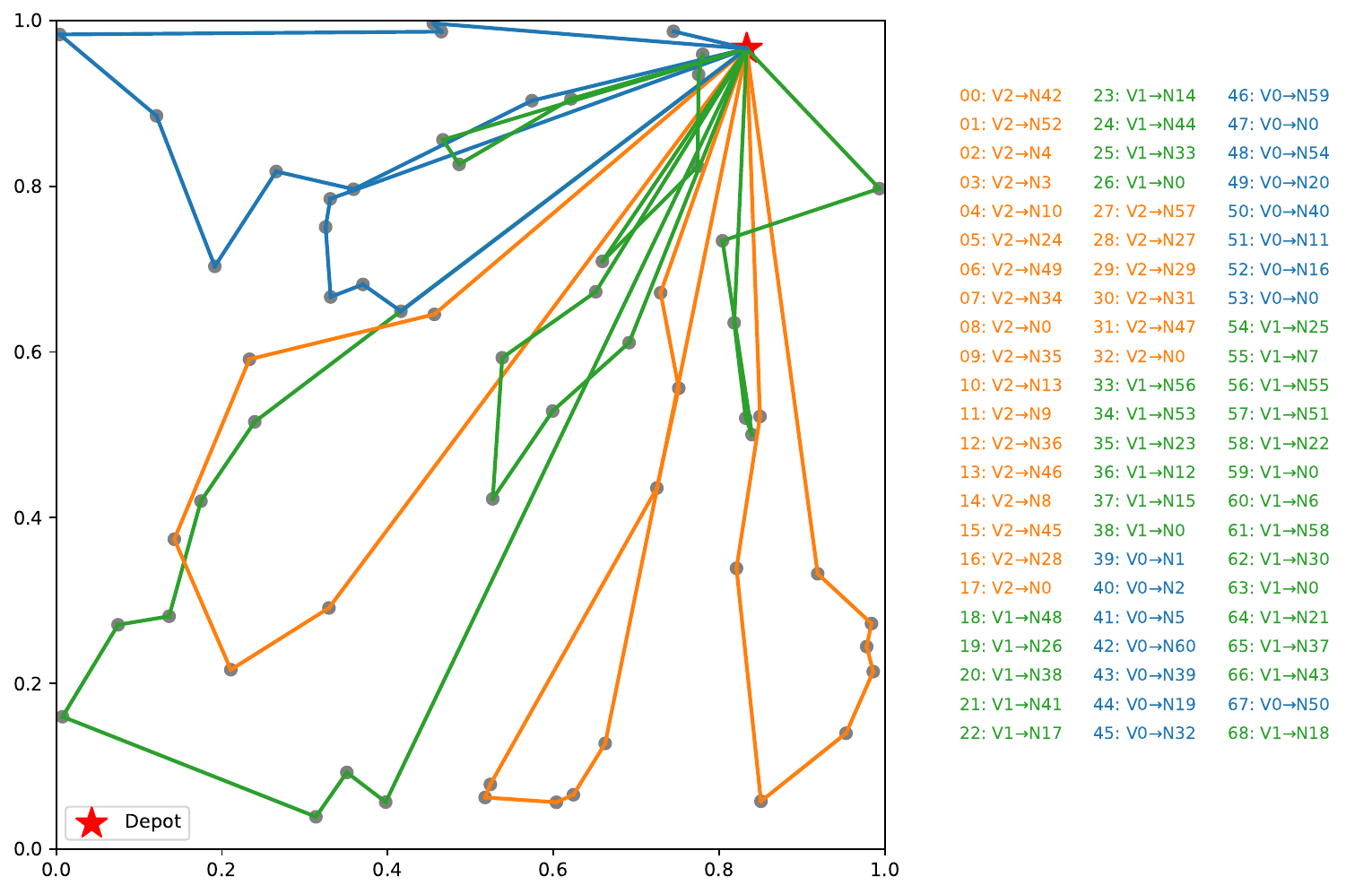}
\caption{ECHO (Obj. 4.79)}
\end{subfigure}
\caption{Comparison of decoding trajectories produced by 2D-Ptr and ECHO.}
\label{vis}
\end{figure}

\section{Discussions}
\label{sec6}
From a macro perspective, ECHO and other  NCO solvers designed for VRPs share a fundamental limitation in that their performance may degrade on real-world instances, because they are predominantly trained on synthetic data with predefined distributions that can be accurately modeled. In contrast, real-world VRP instances often exhibit highly uncertain and complex distribution patterns that are difficult to model accurately. Consequently, even models with strong generalization capabilities may experience inevitable performance degradation when deployed in real-world scenarios \citep{wu_2024_survey}.
Moreover, ECHO assumes Euclidean distance when constructing routing costs, whereas real-world logistics systems typically involve more complex measures, e.g., Manhattan distance in urban grids \citep{son2026towards}. Therefore, future research could focus on developing NCO solvers that incorporate real-world distance and cost metrics, thereby improving their adaptability to practical path planning scenarios.

\section{Conclusions}
\label{sec7}
To better address the practical challenges of MMHCVRP, we propose ECHO, a novel NCO solver. Specifically, ECHO employs our proprietary dual‑modality node encoder to capture local topological relationships among nodes. In addition, by incorporating the proposed PFCA mechanism, ECHO mitigates myopic decisions by emphasizing vehicles selected in the preceding decoding step. Finally, to stabilize the RL training process, ECHO introduces the tailored data augment method that leverages MMHCVRP's inherent properties, i.e., vehicle permutation invariance and node symmetry. The experimental results demonstrate the effectiveness of ECHO across varying numbers of vehicles and nodes, its robust generalization across scales and distribution patterns, and the contribution of each designed methods.

\section{Acknowledgments}
This work is supported in part by the Jilin Provincial Department of Science and Technology Project under Grant 20240101369JC.




\bibliography{hcvrp}
\end{document}